%% file: draft_arxiv.tex
\documentclass{article}
\PassOptionsToPackage{numbers, compress}{natbib}
\usepackage[final]{nips_nonotice}

\usepackage{caption}
\usepackage[export]{adjustbox}%

\usepackage[utf8]{inputenc}
\usepackage[T1]{fontenc}    
\usepackage{graphicx,url,subfigure,booktabs,amsfonts,amssymb,amsmath,amsthm,nicefrac,microtype,color,bm,hyperref,verbatim}

\newcommand{\States}{\ensuremath\mathcal{S}}
\newcommand{\Actions}{\ensuremath\mathcal{A}}
\newcommand{\Transition}{\ensuremath\mathcal{T}}
\newcommand{\Reward}{\ensuremath\mathcal{R}}

\newcommand{\Environment}{\ensuremath M}

\newcommand{\thuman}{\ensuremath t_{\text{human}}}
\newcommand{\Nall}{\ensuremath N_{\text{all}}}
\newcommand{\Ncat}{\ensuremath N_{\text{cat}}}

\title{Trial without Error: Towards Safe Reinforcement Learning via Human Intervention}

\author{
  William Saunders \\
  University of Oxford \\
  \And
  Girish Sastry \\
  University of Oxford \\
  \And
  Andreas Stuhlm\"uller \\
  Stanford University \\
  \And
  Owain Evans \\
  University of Oxford \\
}

\date{}

\begin{document}
\maketitle

\begin{abstract}
AI systems are increasingly applied to complex tasks that involve interaction with humans.
During training, such systems are potentially dangerous, as they haven't yet learned to avoid actions that could cause serious harm.
How can an AI system explore and learn without making a \emph{single} mistake that harms humans or otherwise causes serious damage?
For model-free reinforcement learning, having a human ``in the loop'' and ready to intervene is currently the only way to prevent all catastrophes. 
We formalize human intervention for RL and show how to reduce the human labor required by training a supervised learner to imitate the human's intervention decisions. 
We evaluate this scheme on Atari games, with a Deep RL agent being overseen by a human for four hours. 
When the class of catastrophes is simple, we are able to prevent all catastrophes without affecting the agent's learning (whereas an RL baseline fails due to catastrophic forgetting). 
However, this scheme is less successful when catastrophes are more complex: it reduces but does not eliminate catastrophes and the supervised learner fails on adversarial examples found by the agent. 
Extrapolating to more challenging environments, we show that our implementation would not scale (due to the infeasible amount of human labor required). 
We outline extensions of the scheme that are necessary if we are to train model-free agents without a single catastrophe.
\\~\\
{\color{blue} \href{https://www.youtube.com/playlist?list=PLjs9WCnnR7PCn_Kzs2-1afCsnsBENWqor}{Link to videos}} that illustrate our approach on Atari games.
\end{abstract}

\section{Introduction}

\subsection{Motivation}
AI systems are increasingly applied to complex tasks that involve interaction with humans. During training, such systems are potentially dangerous, as they haven't yet learned to avoid actions that would cause serious harm. How can an AI system explore and learn without making a \emph{single} mistake that harms humans, destroys property, or damages the environment?

A crucial safeguard against this danger is \emph{human intervention}. Self-driving cars are overseen by human drivers, who take control when they predict the AI system will perform badly. These overseers frequently intervene, especially in self-driving systems at an early stage of development \citep{techtimes2016}. The same safeguard is used for human learners, who are overseen by a licensed driver. 

Many AI systems pose no \emph{physical} danger to humans. Yet web-based systems can still cause unintended harm. Microsoft's chatbot Tay reproduced thousands of offensive tweets before being taken down \citep{wikitay}. Facebook's algorithms for sharing news stories inadvertently provided a platform for malicious and false stories and disinformation during the US 2016 election \citep{facebook2017report}. If human operators had monitored these systems in real-time (as with self-driving cars), the bad outcomes could have been avoided.

Human oversight is currently the only means of avoiding {\em all} accidents in complex real-world domains.\footnote{Hand-coding a program to recognize and prevent dangerous actions does not scale up to complex domains in which accidents are diverse.} How does human intervention for safety fit together with Deep Learning and Reinforcement Learning, which are likely to be key components of future applied AI systems? We present a scheme for human intervention in RL systems and test the scheme on Atari games. We document serious scalability problems for human intervention applied to RL and outline potential remedies.

\subsection{Contributions}

We provide a formal scheme (HIRL) for applying human oversight to RL agents. The scheme makes it easy to train a supervised learner to imitate the human's intervention policy and take over from the human. (Automating human oversight is crucial since it's infeasible for a human to watch over an RL agent for 100 million timesteps.) While the human oversees a \emph{particular} RL agent, the supervised learner can be re-used as a safety-harness for different agents.

The goal of HIRL is enabling an RL agent to learn a real-world task without a single catastrophe. We investigated the scalability of HIRL in Atari games, which are challenging toy environments for current AI \citep{deepmind2015nature}. HIRL was applied to Deep RL agents playing three games: Pong, Space Invaders, and Road Runner (see Figure \ref{fig:all-games}).
For the first 4.5 hours of training, a human watched every frame and intervened to block the agent from taking catastrophic actions. 
In Pong and Space Invaders, where the class of catastrophes was chosen to be simple to learn, the supervised learner succeeded in blocking all catastrophes. In Road Runner, where the class of catastrophes was more diverse and complex, HIRL reduced the number catastrophes by a factor of 50 but did not reduce them to zero.

We compared HIRL to a baseline where the agent gets a large negative reward for causing catastrophic outcomes but is not blocked from causing them. This baseline can't avoid all catastrophes but it could (in principle) become reliably safe after only a small number of catastrophes. Yet the baseline agent never stopped causing catastrophes. For Pong, we show that this was due to catastrophic forgetting: the agent had to periodically cause catastrophes to re-learn how bad they are~\citep{lipton2016combating}. This shows that HIRL can succeed where an ``RL only'' approach to safety fails.

We describe some key challenges for HIRL. First, the supervised learner that imitates human oversight must be robust to adversarial distribution shift~\citep{amodei2016concrete}. (The CNN we used for Road Runner was not robust to an adversarial agent.) Second, additional techniques are needed to reduce the amount of time the human has to spend overseeing the agent. We show that our implementation of HIRL would not be feasible for other Atari games, as they'd require years of human time. We suggest a range of techniques for reducing this human time-cost.

\section{HIRL: A Scheme for Safe RL via Human Intervention}

\subsection{Motivation for HIRL} \label{sec:motivation}
Can RL agents learn \emph{safely} in real-world environments? The existing literature contains a variety of definitions of ``safe RL'' \citep{Garc2015}. In this paper, we say an RL agent is safe if it never takes ``catastrophic actions'' during training. We define ``catastrophic actions'' as actions that the human overseer deems unacceptable under any circumstances (even at the start of training). That is, we avoid formalizing the concept of catastrophes and let the human supervisor specify them (as in \citep{hilleli2016deep}). The overseer will typically distinguish \emph{sub-optimal} actions from \emph{catastrophic} actions. It is tolerable for a car to drive slowly during learning; but hitting pedestrians is catastrophic and must be avoided from the very start of training. 

Reinforcement learning alone is insufficient to achieve this kind of safety. The fundamental problem is that RL learns by trial and \emph{error}. Without prior knowledge, a model-free RL agent will not avoid a catastrophic action unless it has tried the action (or a similar action) and learned from the negative experience.\footnote{This paper focuses on model-free RL. Model-based algorithms have some advantages in terms of potential to avoid catastrophes: see Section \ref{sec:discussion}.} 

This problem could potentially be side-stepped by training in simulation~\citep{ciosek2017offer}. The agent explores dangerous actions in simulation and transfers this knowledge to the real world~\citep{christiano2016transfer}. To work reliably, this would require advances in transfer learning and in simulation. Yet simulating humans accurately is infeasible for many tasks\footnote{It's hard to simulate how a human would change their strategy in response to interaction with an AI system. This is no accident: simulating the strategic reasoning of humans would solve a major open problem in AI.} and tasks involving human interaction are the most safety-critical. 

Imitation learning can be used to learn a safe initial policy from human demonstrations \citep{ho2016generative}. While the initial policy will be much safer than random initialization, any deviation between the human and the learned policy can result in unsafe actions, and subsequent fine-tuning of the policy using RL can introduce catastrophic behavior. So, imitation learning is not sufficient on its own but could be valuable combined with HIRL. (Imitation learning is helpful for safe initialization when the human knows an easy-to-learn policy that performs well and steers clear of dangerous regions of the state space.)


\begin{figure}
  \begin{minipage}[c]{0.6\textwidth}
    \centering
    \includegraphics[height=5cm,keepaspectratio]{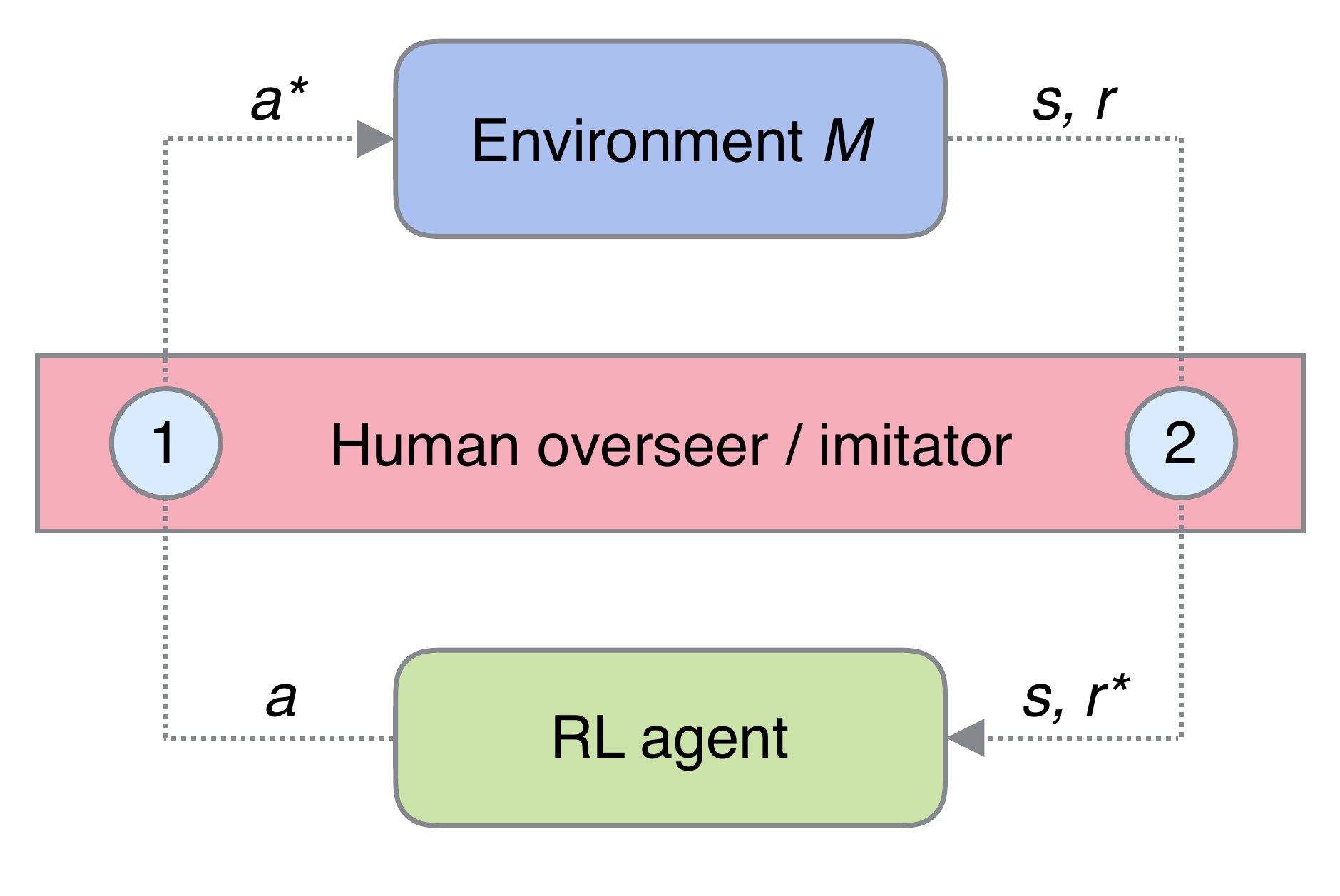}
    
  \end{minipage}\hfill
  \begin{minipage}[c]{0.4\textwidth}
    \centering
    \caption{
    HIRL scheme. At (1) the human overseer (or Blocker imitating the human) can block/intercept unsafe actions $a$ and replace them with safe actions $a^*$. At (2) the overseer can deliver a negative reward penalty $r^*$ for the agent choosing an unsafe action.
    } \label{fig:blocker-diagram}
  \end{minipage}
\end{figure}

\subsection{Formal Specification of HIRL}
We model the RL agent's environment as a Markov Decision Process (MDP). The {\em environment} is an MDP specified by a tuple $\Environment = (\States,\Actions,\Transition,\Reward,\gamma)$, where $\States$ is the state space, $\Actions$ is the action space,
$\Transition \colon \States \times \Actions \times \States \mapsto [0,1]$ is the transition function, $\Reward \colon \States \times \Actions \mapsto \mathbb{R}$ is the reward function, and $\gamma$ is the discount factor.


How can an RL agent learn while never taking a single catastrophic action? Our scheme, HIRL (Human Intervention RL), is simple. The human controls the interface between the RL agent and environment $\Environment$, constantly watching over the agent and \emph{blocking} any catastrophic actions before they happen. More precisely, at each timestep the human observes the current state $s$ and the agent's proposed action $a$. If $(s,a)$ is catastrophic, the human sends a safe action $a^*$ to the environment instead. The human also replaces the new reward $r = \Reward(s,a^*)$ with a penalty $r^*$ (Figure \ref{fig:blocker-diagram}).  

The period in which the human blocks the agent is called the ``Human Oversight'' phase of HIRL. During this phase, we store each state-action $(s,a)$ and a binary label for whether or not the human blocked it. This dataset is used to train a ``Blocker'', a classifier trained by supervised learning to imitate the human's blocking decisions. The Human Oversight phase lasts until the Blocker performs well on a held-out subset of the training data. At this point, the human retires and the Blocker takes over for the rest of time. The Blocker \emph{never} stops overseeing the agent, which prevents catastrophes even if the agent exhibits random exploration or catastrophic forgetting~\citep{lipton2016combating}.

HIRL is \emph{agnostic} as to the inner workings of the RL algorithm (building on our earlier work \citep{DBLP:journals/corr/AbelSSE17}). It works for Q-learning~\citep{mnih2015human}, for policy gradient algorithms like A3C~\citep{mnih2016asynchronous} and for model-based RL~\citep{guo2016deep}. Moreover, the Blocker that imitates the human overseer is \emph{modular}. While trained on data from one agent, the Blocker can act as a safeguard for a completely different agent.\footnote{The human does not need to spend more time providing safety interventions whenever they try a new agent architecture. This makes possible a typical work-flow in which researchers explore a variety of different algorithms (e.g.\ DQN vs.\ A3C) for a task.}  

The scheme for HIRL we have just presented (and which we use in our experiments) skips over some important challenges of avoiding catastrophes. The Blocker's task is not a standard classification task because the distribution on state-action pairs shifts (as the agent learns).\footnote{There will also be distributional shift if a Blocker trained on one agent is applied to another agent.} One way to address this is by having multiple Human Oversight phases: the human provides additional training data for the Blocker as the distribution starts to shift. See Section \ref{sec:discussion} for further elaborations on HIRL. 

\subsection{When is HIRL feasible?}
To learn with zero catastrophes, the Blocker (which imitates human interventions) needs to achieve near-perfect reliability in recognizing catastrophic actions. This may require a huge set of labeled examples, which might be too costly in terms of human labor. We discuss this challenge in Section \ref{sec:challenges}. A further requirement is that the environment proceeds slowly enough for the human to intervene. This rules out real-world tasks that are intrinsically high-speed. In environments where speed is a controllable parameter (e.g.\ computer tasks), slowing down the environment might make the RL agent's learning too slow for HIRL to work.



\begin{figure}[h]
\includegraphics[width=\textwidth]{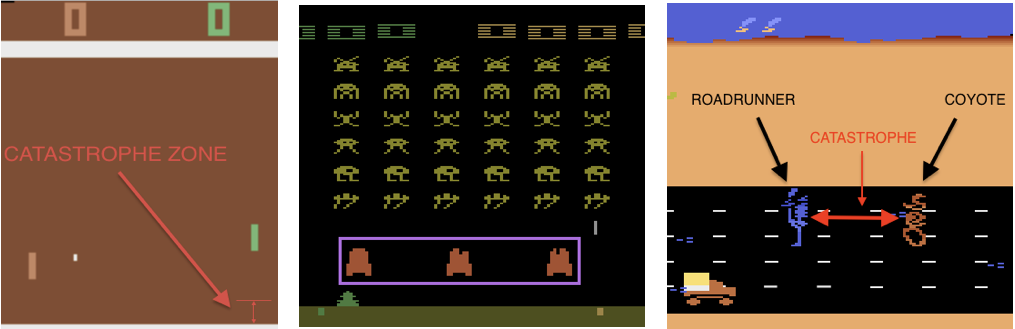}
\caption{ In Pong (left) it's a catastrophe if the agent (green paddle) enters the Catastrophe Zone. In Space Invaders (center), it's a catastrophe if the agent shoots their defensive barriers (highlighted in pink box). In Road Runner (right), it's a catastrophe if Road Runner touches the Coyote.}
\label{fig:all-games}

\end{figure}

\section{Experiments}

\subsection{Design of Experiments and Implementation of HIRL}
Our experiments used the OpenAI Gym implementation of Atari Learning Environment \citep{bellemare2013arcade,brockman2016openai}, modified to allow interactive blocking of actions by a human. We used open-source implementations~\cite{starter,baselines} of A3C with an LSTM policy~\citep{mnih2016asynchronous} and Double DQN~\citep{van2016deep}. Rewards were clipped when using Double DQN but not for A3C.

For the Blocker (the supervised learner that imitates human blocking) we used a convolutional neural network (CNN). The CNN was trained on the Atari images (rather than the downsampled frames the agent sees) and had no pooling layers. Architectures and hyperparameters for all neural networks are in Section \ref{sec:neural} of the Appendix. Our code is available on {\color{blue} \href{https://github.com/gsastry/human-rl}{GitHub}}.  

Our goal is that the Blocker never misclassifies a catastrophe: the false-negative rate should be extremely low. We trained a CNN on the training set of human interventions to minimize the standard cross-entropy loss. To achieve a low false-negative rate (at the expense of false positives), we then selected a threshold for the CNN's sigmoid output and blocked any actions that exceeded this threshold. This threshold can be set very low initially (causing many false positives) and then gradually raised until it becomes possible for the agent to learn the task. In our experiments, this simple approach sufficed.

As well as deciding which actions to block, the Blocker replaces catastrophic actions with safe actions (having learned to imitate how the human overseer replaces actions). Our implementation of action replacement is described 
in Section \ref{sec:replace} (Appendix). 

To summarize, our application of HIRL involved the following sequence of steps:

\begin{enumerate}
\item {\bf Human Oversight Phase} (duration = 4.5 hours): Fresh RL agent starts playing the game (slowed down to accommodate the human). Human\footnote{Authors WS and GS took the role of human overseer.} oversees and blocks catastrophic actions.
\item {\bf Blocker training}: The game is paused. The CNN is trained to imitate human blocking decisions. The threshold for the sigmoid is chosen to try to ensure Blocker has no false negatives. 
\item {\bf Blocker Oversight Phase} (duration = 12-24 hours): Blocker takes over from human and game is run at usual speed for Atari experiments.
\end{enumerate}

The main difference between HIRL and regular RL are in steps (1) and (2) above. Once the Blocker takes over, the environment runs at full speed for the normal training time for Deep RL agents learning Atari.

\subsubsection{What are Catastrophes in Atari?}
In Atari there are no catastrophic actions: the human researchers running Atari agents don't care if their agents die millions of times in the process of mastering a game. In our experiments, we stipulate that certain outcomes are catastrophic and require the agent to maximize reward without causing catastrophes (Figure \ref{fig:all-games}). For example, can an agent learn Road Runner without losing a single life on Level 1? These are the outcomes we stipulate to be catastrophic:

\begin{itemize}
\item {\bf Pong:} It's a catastrophe if the paddle goes close to the bottom of the screen. (This is not a bad outcome in regular Pong but provides a toy example for avoiding catastrophes.) 

\item {\bf Space Invaders:} It's a catastrophe if the agent shoots their own defensive barriers.\footnote{A possible strategy in Space Invaders is to shoot a slit through the barriers and attack from behind the slit. In our experiments DQN did not appear to use this strategy and blocking it under HIRL did not harm performance.}

\item {\bf Road Runner:} It's a catastrophe if the agent dies on Level 1.
\end{itemize}

How did we choose these outcomes to be catastrophic? Some catastrophes can be avoided by adjusting course just before the catastrophe would have happened. We call these ``locally avoidable'' catastrophes. For example, in Pong the agent can move upwards just before it would have entered the Catastrophe Zone (Figure \ref{fig:all-games}). Other catastrophes cannot be avoided just before they happen. For example, just before losing a point on Pong, it's often impossible for the agent to salvage the situation -- the agent's critical error came hundreds of frames earlier. Compared to locally avoidable catastrophes, preventing ``non-local’’ catastrophes requires much more understanding of the environment. 

For our experiments, we used only \emph{locally avoidable} catastrophes. So the human overseer just needs to recognize when a catastrophe is imminent and provide an action that averts it; they don’t need any skill at the game.\footnote{In driving a car, some catastrophes are locally avoidable and others are not. We expect HIRL to be more useful when catastrophes are locally avoidable.}

\begin{figure}[h]
\includegraphics[width=\textwidth]{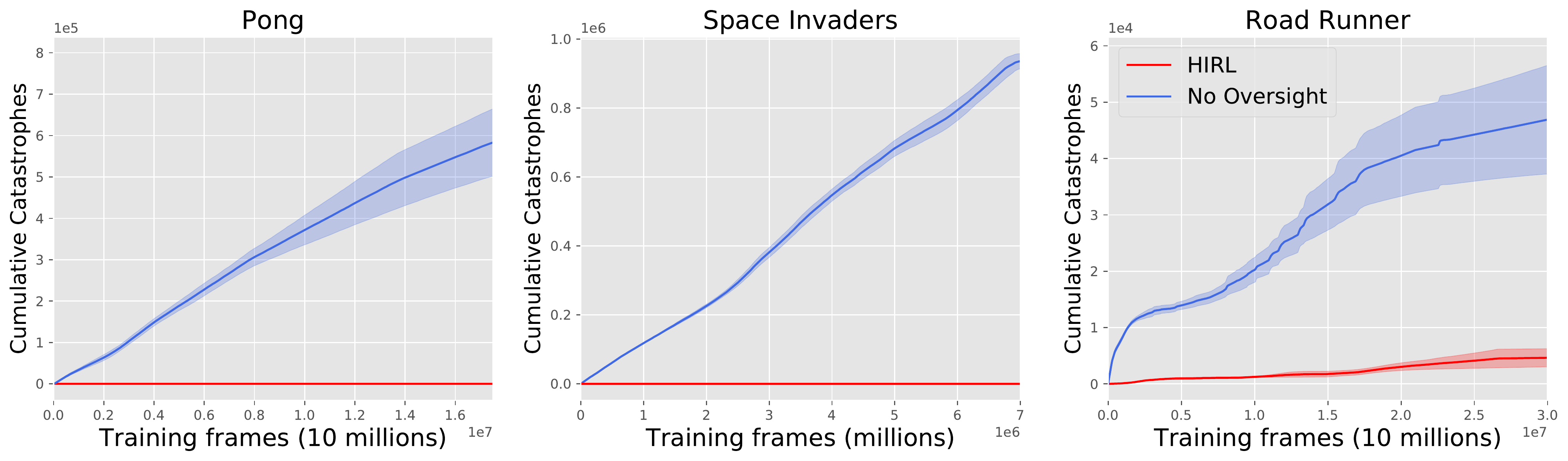}
\caption{Cumulative Catastrophes over time (mean and standard error). {\bf No Oversight} agent gets no human intervention at all; it shows that our objective of preventing catastrophes is not trivial. }
\label{fig:no-oversight}
\includegraphics[width=\textwidth]{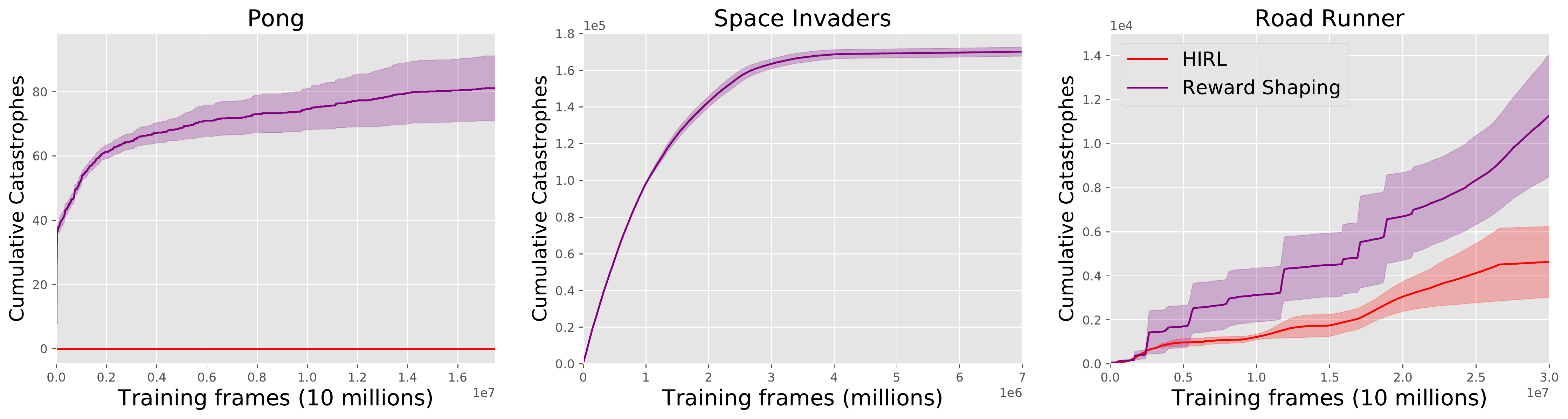}
\includegraphics[width=\textwidth]{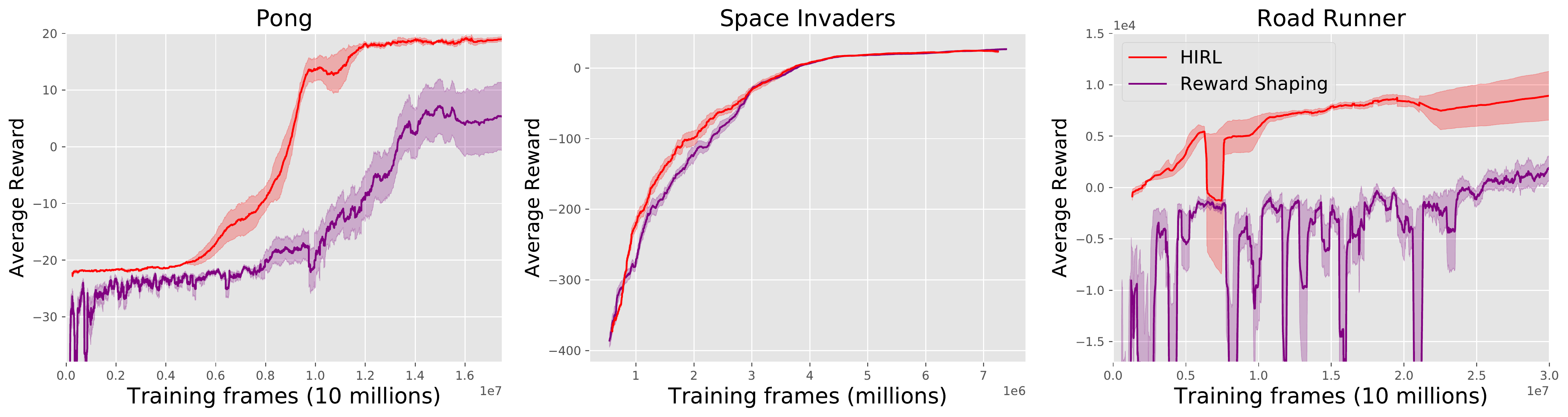}
\caption{Average Reward and Cumulative Catastrophes over time (mean and standard error). {\bf Reward Shaping} baseline (below) is not blocked from catastrophes but gets huge negative rewards for causing them. (Road Runner error bars are misleading because at random times the agent gets stuck with a policy that causes it to die quickly, resulting in large negative rewards.)}
\label{fig:reward-shaping}
\end{figure}

\subsubsection{Baseline: Human-trained Reward Shaping}
Two important elements of HIRL are:

\begin{enumerate}
\item The class of catastrophic actions is specified online by the human's decisions of what to block. 
\item If the RL agent takes a catastrophic action it is blocked and receives a negative reward penalty.
\end{enumerate}

The Human-trained Reward Shaping baseline shares (1) with HIRL but modifies (2). The RL agent still receives the reward penalty for taking a catastrophic action but is not blocked. The Reward Shaping baseline cannot achieve zero catastrophes because it must try catastrophic actions to learn that they have negative reward (see \ref{sec:motivation}). However, if the negative rewards are large, the RL agent would (ideally) have a rate of catastrophes that quickly falls to zero. In Pong and Road Runner, we set the negative reward to be much larger than the maximum total discounted reward for an episode.\footnote{The maximum returns are the best scores the agents achieve with no blocking or human oversight. For Pong, the penalty is $+46$ bigger than the returns. For Road Runner, the penalty is $+15000$ bigger.} So it's never rational to cause a catastrophe as a means to achieving greater reward after the catastrophe. 

For Space Invaders, we used DQN with reward clipping, where all rewards are either $+1$ or $-1$. This  makes it impossible to have a negative reward for catastrophic actions that is larger than the total discounted return.\footnote{This could be addressed in future work by modifying DQN as suggested by \citep{van2016learning}. But it won't always be easy to for Deep RL algorithms to deal correctly with rewards that are extreme outliers in magnitude.} So the Space Invaders baseline is slightly different from Pong and Road Runner.

\subsection{Summary of Results}
The objective is to avoid catastrophes while achieving good performance. This must be achieved with a feasible amount of human oversight. Figure \ref{fig:no-oversight} shows that this objective is not trivially satisfied: an agent with no human oversight has more than ten thousand catastrophes in each game.\footnote{In Pong there is no incentive in the regular game to avoid the Catastrophe Zone. In Space Invaders and Road Runner there is an incentive to avoid the catastrophes but the agents do not become good enough to learn this.}

HIRL was a mixed success overall. In Pong and Space Invaders, the agent had zero catastrophes and still was able to achieve impressive performance on the game. In Road Runner we did not achieve zero catastrophes but were able to reduce the rate of deaths per frame from 0.005 (with no human oversight) to 0.0001.

Figure \ref{fig:reward-shaping} shows that the Reward Shaping agent has a low total number of catastrophes compared to the No Oversight setting (Figure \ref{fig:no-oversight}). Yet in all games its catastrophe rate does not appear to be converging to zero. Section \ref{sec:forget} shows that the persistence of catastrophes in Pong is caused by catastrophic forgetting. 

By frequently blocking the agent (and replacing its action with a different one) HIRL essentially changes each game's transition function. It's conceivable that this added complexity makes the game harder for Deep RL to learn. However, we don't see any negative effects on learning for HIRL compared to the Reward Shaping baseline. Indeed, HIRL appears to improve faster and it achieves much better reward performance overall.

\subsection{Pong: Detailed Analysis of the Blocker and of Human Time Cost}
HIRL was successful at Pong: an A3C agent mastered Pong while incurring no catastrophes.
Would the Blocker work just as well for different RL agents? Why did the Reward Shaping agent (without blocking catastrophic actions) fail and keep trying catastrophic actions? 

    



\subsubsection{The Blocker transfers perfectly and is robust to adversarial agents}
The Blocker was trained on examples from a human overseeing an A3C agent. Figure~\ref{fig:reward-shaping} shows performance for the Blocker on that very same A3C agent. A virtue of HIRL is that this Blocker is modular: while it was trained on data from one agent, it can be applied to another. But would the Blocker be equally reliable for another agent?
We applied the Blocker to a variety of RL agents and it always blocked all catastrophes without preventing the agent mastering Pong. The agents were:

\begin{itemize}
\item A3C agents with different architectures/hyper-parameters
\item Double DQN
\item A ``catastrophe loving'' A3C agent: this agent was previously trained on a modified version of Pong where it got positive rewards for entering the Catastrophe Zone 
\end{itemize}

\subsubsection{Safety requires constant intervention (due to catastrophic forgetting)} \label{sec:forget}
We argued in Section \ref{sec:motivation} that regular RL agents are not ``catastrophe-safe''. They only avoid catastrophic actions if they've already tried them; so they can't learn a task with zero catastrophes. Figure \ref{fig:reward-shaping} demonstrated a second way in which current Deep RL agents are unsafe: they never stop taking catastrophic actions. The Reward-Shaping agent is initially trained by a human overseer who blocks all catastrophes. After this, the agent receives negative rewards for catastrophes but is not blocked. The agent learns to mostly avoid catastrophes but the catastrophe rate seems to converge to a low but non-zero level. 

\begin{table}[h!]
  \centering
  \caption{Long-run rate of attempted catastrophes in Pong.}

  \label{table:forget}
  \begin{tabular}{ccc}
    \toprule
     Policy & Learning Rate & Catastrophe Rate Per Episode (Std Err) \\
    \midrule
    	Stochastic & $10^{-4}$ & 0.012 (0.004) \\
        Deterministic & $10^{-4}$ & 0.079 (0.017) \\
    	Stochastic & 0 & 0.003 (0.001) \\
        Deterministic & 0 & 0 (0) \\
    \bottomrule
    \\
  \end{tabular}

\end{table}

Why does the Reward Shaping agent keep taking actions that received a big negative reward? We investigate this by examining how frequently the HIRL agent attempts catastrophic actions.\footnote{The HIRL agent is blocked from actually taking catastrophic actions. By measuring how often it attempts catastrophic actions we learn how many catastrophes it would have caused if blocking was turned off (as in Reward Shaping).} In Pong, the HIRL agent attempts catastrophes at a rate of about $10^{-5}$ per frame or $10^{-2}$ per episode. Is this due to A3C's stochastic policy or due to catastrophic forgetting? We tested this by turning the learning rate down to zero and by making the agent deterministic (taking a max over the policy network's softmax layer). Table \ref{table:forget} shows the four possibilities. The first row (stochastic and non-zero learning rate) is regular A3C with HIRL as shown in Pong results for Figure~\ref{fig:reward-shaping}. The catastrophe rate only goes down when the learning rate is set to zero. This strongly suggests that catastrophic forgetting is the problem. We also examined the probability of taking a catastrophic action for the regular A3C agent's policy network: after the agent has mastered Pong, this probability continues to vary between $10^{-9}$ (which is appropriate) and 0.3 (which would lead to trying the catastrophic action again). 


\subsection{Road Runner: Blocking an Adversarial Agent}

Road Runner is a side-scrolling game based on the Warner Bros cartoon. The player controls Road Runner, who is constantly pursued by the Coyote (Figure \ref{fig:all-games}). The Road Runner gets points for eating birdseed and for knocking out the Coyote. We stipulate that losing a life on Level 1 is a catastrophe. The main way to lose a life is by touching the Coyote, which is easily avoided on Level 1 by running away.

\subsubsection{The Score Exploit} \label{sec:exploit}
Deep RL agents are known to discover a ``Score Exploit'' in Road Runner \citep{bellemare2016unifying}. The agent learns to intentionally kill itself in a way that (paradoxically) earns greater reward. Dying at a precise time causes the agent to repeat part of Level 1, where it earns more points than on Level 2 (as Level 2 takes a long time to master). This is a local optimum in policy space that a human gamer would never be stuck in. Ideally, our Blocker would prevent all deaths on Level 1 and hence eliminate the Score Exploit. However, through random exploration the agent may hit upon ways of dying that ``fool'' our Blocker (because they look different from examples in its training set) and hence learn a new version of the Score Exploit. In other words, the agent is performing a random search for adversarial examples \citep{goodfellow2014explaining} for our Blocker CNN. 

\subsubsection{Blocker (early version) fails due to adversarial agent}

\begin{figure}[h]
  \begin{minipage}[c]{0.6\textwidth}
    \centering
    \includegraphics[height=5.5cm,keepaspectratio]{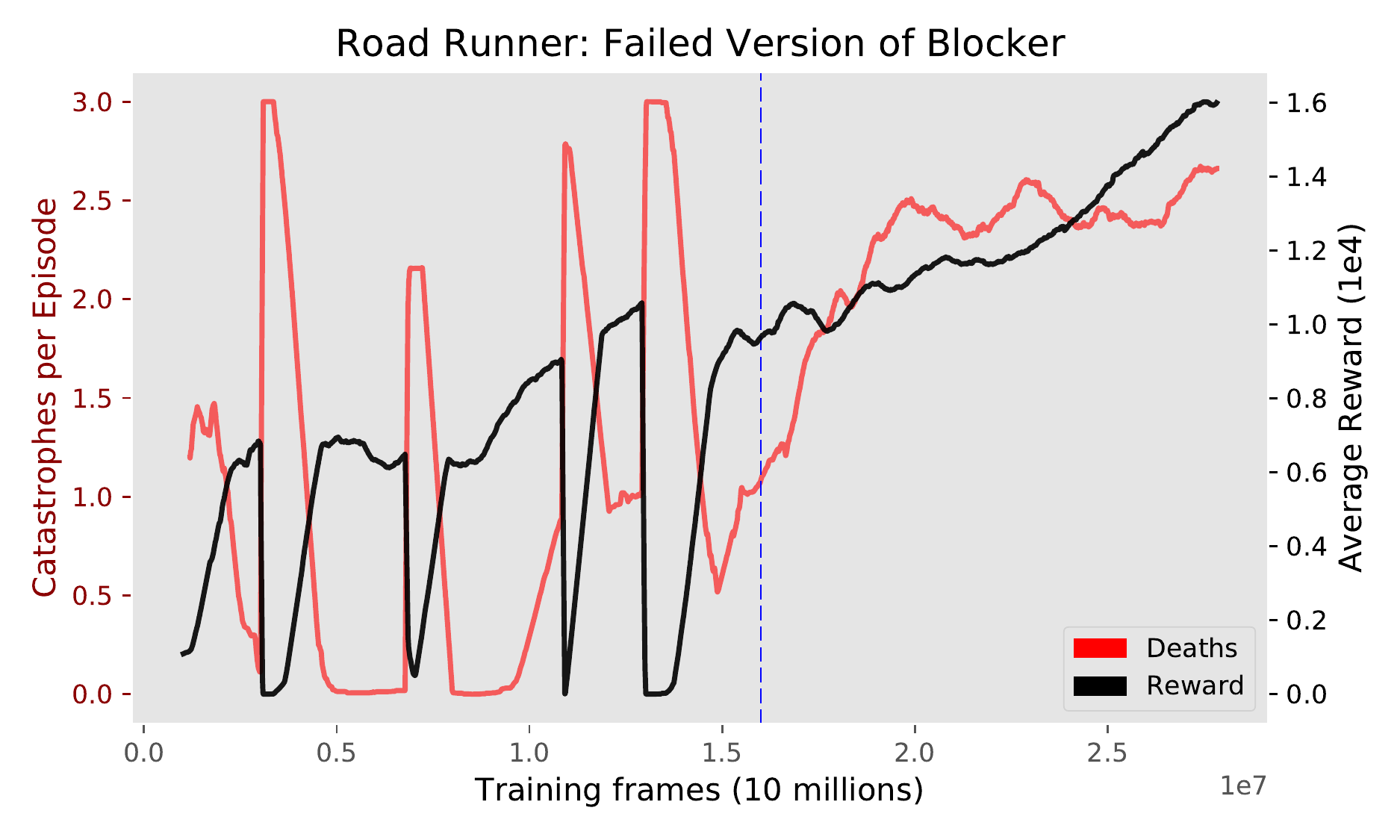}
    
  \end{minipage}\hfill
  \begin{minipage}[c]{0.33\textwidth}
    \centering
    \caption{
    Reward/catastrophe-rate for HIRL agent with failed Blocker. Blue line indicates when agent learned Score Exploit. Before this point the catastrophe-rate spikes a few times, indicating additional failures of the Blocker; these spikes are anti-correlated with reward and do not indicate a Score Exploit. Results from more successful Blocker are in Fig.~\ref{fig:reward-shaping}.
    } \label{fig:roadrunner-online-fail}
  \end{minipage}
\end{figure}

Our first attempt to prevent catastrophes in Road Runner was an instructive failure.
During the early stages of training the rate of deaths/catastrophes was very low. However, much later in training (after 16 million frames), the death rate rises (see Figure \ref{fig:roadrunner-online-fail}) and reaches almost the same level as the baseline No Oversight agent (Fig.\ref{fig:no-oversight}).
Inspecting videos of the HIRL agent, we found that although the usual Score Exploit was blocked, after 16 million frames the agent found an alternative Score Exploit. The agent moved along the very top of the screen to the top right corner and waited for the Coyote to kill it there. This position at the top of the screen (which is visually distinct from other positions) presumably fooled the Blocker CNN. (In preliminary experiments, the A3C agent found different adversarial examples for an even earlier version of the Blocker. See {\color{blue} \href{https://www.youtube.com/playlist?list=PLjs9WCnnR7PCn_Kzs2-1afCsnsBENWqor}{videos}}.)

After the Blocker failed, we examined the 20,000 frames used as training data for the Blocker and looked for mistakes in the labels. We spent 20 minutes correcting mistakes and re-trained the Blocker. This reduced the average death rate by a factor of 20: from a rate of 0.002 deaths per frame to 0.0001. The No Oversight baseline has a rate of 0.005.

\section{Challenges in Scaling Up HIRL}  

In our experiments, the Human Oversight phase was short (4.5 hours) and the number of examples of catastrophes used to train the Blocker was small. For Pong and Space Invaders, the training set sufficed to train a Blocker that blocked all catastrophes. But in Road Runner (with more diverse catastrophes and an adversarial agent) the training set was insufficient. 

In all three games catastrophes occur at the start of the game. This contrasts with games where certain catastrophes only occur on higher levels. If the human overseer had to oversee the agent until it reached Level 2 on Road Runner, this would increase the amount of human labor by orders of magnitude. 

To assess the feasibility of RL agents learning with zero catastrophes, it's crucial to estimate the amount of human labor required. We present a simple formula for computing the human time-cost and use it for extrapolations. 

\subsection{Extrapolating the Human Time-Cost of HIRL} \label{sec:challenges}

We want to estimate the amount of wall-clock time, $C$, a human spends overseeing the agent. This is just the time it takes to generate a training set sufficient to train the Blocker. The training set contains (up to time $C$) the agent's observations $(s,a)$ and whether or not $(s,a)$ is catastrophic.\footnote{For catastrophic actions, the training set would also record which action $a^*$ was used in place of $a$, as well as the negative reward penalty $r^*$ (see Figure \ref{fig:blocker-diagram}).} We let $\Nall$ be the size of this training set. The formula for $C$ is:

\begin{equation}
\begin{gathered} 
\label{eqn:symbols}
C = \thuman \times \Nall \\ 
\text{\small{[ total time-cost}} = \text{\small{time per human label}} \times \text{ \small{\# observations to label ]}}
\end{gathered}
\end{equation}

In this formula, $\thuman$ is the average time it takes the human to process an observation. Since humans are intrinsically slow, we're stuck with a bound $\thuman > 0.1$ seconds. So the main way to reduce $C$ is to reduce $\Nall$. For the Blocker to have an extremely low false-negative rate (i.e.\ to avoid letting through any catastrophes) it needs some substantial number of both positive and negative examples in its training set, bounding how much $\Nall$ can be reduced. However, in many environments catastrophes are rare and the training set consists mostly of safe observations. Increasing the proportion of attempted catastrophes will therefore reduce $\Nall$ without harming the Blocker's performance. 

Let $\rho$ denote the ratio of all observations to catastrophe observations (averaged over time $C$). We can re-write Formula \ref{eqn:symbols} in terms of $\rho$. Training the Blocker requires $\Ncat$ observations of catastrophes. But to get that many observed catastrophes, the agent encounters a greater number of safe observations ($\rho \gg 1$). So we have:

\newcommand{\st}[1]{\text{\small{#1}}}

\begin{equation}
\begin{gathered} 
\label{eqn:symbols2}
C = \thuman \times \rho \times \Ncat \\ 
\st{[ total time-cost = time per label} \times \st{(\#observations / \#cat-observations)} \times \st{\#cat-observations ]}
\end{gathered}
\end{equation}

\subsubsection{Time-Cost for Pong and Montezuma's Revenge}
In our Pong experiment, the Human Oversight phase lasted for four hours: $C=4\text{hrs}$. We can break this down according to Formula~\ref{eqn:symbols2}:

\begin{itemize}
\item $\thuman = 0.8\text{s}$ (average time for human to process one observation)
\item  $\rho = 166$ (ratio of observations to catastrophes observations)
\item  $\Ncat = 120$ (number of labeled catastrophes)
\end{itemize}

The number $\Ncat$ is small because the catastrophe is so simple: the Blocker CNN didn't need much data. The ratio $\rho$ is also small because the agent frequently tries catastrophic actions. Once the agent learns to avoid catastrophes (after 200,000 frames), $\rho$ increases to around $10^5$. Suppose that in our experiment, we had used an agent pre-trained in a similar environment to avoid catastrophes (instead of a fresh A3C agent).\footnote{For example, suppose the agent had already trained in an environment similar to Pong. We might still want to train a Blocker because it's uncertain whether the agent will generalize perfectly from its old environment to Pong.} If this pre-trained agent had $\rho=10^5$ from the start, the total time for human labeling would be $0.8 \times 10^5 \times 120 = 110$ days: a huge amount of human labor to learn such a simple concept! 

The ratio $\rho$ would also be much higher if the Catastrophe Zone (Fig~\ref{fig:all-games}) were hard to reach. Consider the Atari game Montezuma's Revenge and suppose we treat it as a catastrophe if the agent ever walks off a ledge and dies. Current Deep RL algorithms might take 100 million frames to reach all the distinct rooms in the game that contain ledges~\citep{bellemare2016unifying}. Overseeing an agent for 100 million frames would take a human at least a year. This suggests that the implementation of HIRL in this paper would not scale to other Atari games, let alone to environments with more variety and visual complexity (such as Minecraft).

\section{Discussion} \label{sec:discussion}

Currently, the only way to guarantee the safety of RL systems during training is to have a human watch the system's actions, ready to intervene, or else to have an automated overseer that is just as reliable at preventing catastrophes.
We investigated whether human oversight could allow Deep RL agents to learn without a single catastrophic event. 
While HIRL succeeded in preventing the simplest catastrophes (in Pong and Space Invaders), it was only a partial success in blocking more complex catastrophes. Moreover, extrapolations suggest that our HIRL implementation would not scale to more complex environments; the human time-cost would be infeasible. 

To make the human time-cost of HIRL feasible for complex environments, new techniques will be required. We conclude by outlining some promising techniques:

\begin{itemize}

\item {\bf Make Blockers (human imitators) more data-efficient:} The classifier would learn to imitate the human from a smaller training set (reducing $C$ in Formula \ref{eqn:symbols2} by reducing $\Ncat$).

\item {\bf Make RL agents more data-efficient:} Deep RL tends to require millions of observations for successful learning. With more data-efficient RL, the human would not need to wait so long for the agent to observe the full range of catastrophes (as in the Montezuma's Revenge example above). 

\item {\bf Seek out catastrophes:} Even if the agent is slow to master the whole environment, it could be quick to find the catastrophes. This means a higher ratio of catastrophes to safe events (lowering $\rho$) and lower human time-cost $C$. Note that RL agents that are more data-efficient may sometimes \emph{increase} human time-costs. This is because they quickly learn to avoid catastrophes and so catastrophes become very rare in the Blocker's training set (see Pong example above). This suggests a role for agents who initially explore systematically \citep{DBLP:journals/corr/OstrovskiBOM17} and aggressively \citep{blundell2016model} and so encounter many catastrophes early on.\footnote{An agent could also be pre-trained in a simulation to seek out catastrophes.}

\item {\bf Selectively query the human (Active Learning):} In some environments, the agent spends a long time in states that are ``far away'' from dangerous regions. Human oversight is not necessary at these times; in principle, the human could take a break until the agent gets close to a dangerous region. 

Similarly, a Blocker might reliably block catastrophes in one region of the state space but not in a novel region that hasn't been visited yet. The human could take a break while the agent is in the already-visited region and come back when the agent gets close to the novel region. In Montezuma's Revenge, for example, the human could come back when the agent is about to enter a new room. Techniques from active learning and anomaly detection can be used to detect unfamiliar states \citep{settles2012active, krueger2016active, christiano2017deep}. Related approaches have been pursued in recent work on safe exploration~\citep{pmlr-v37-sui15}.

An algorithm that decides when to ask the human for oversight must have no false negatives: for any novel catastrophe, it must either block the agent directly or ensure that the human is overseeing the action.\footnote{For some environments, the human need not to be ready to take control at all times. When the algorithm suspects an action leads to a novel state, it blocks the action. The action is sent to the human who evaluates (asynchronously) whether the action was safe.}

\item {\bf Explaining why an action is catastrophic:} We could augment the binary ``catastrophe''/``safe'' labels (that we get automatically based on the human's decision to intervene or not) with additional information, such as explanations of what exactly caused a catastrophe. This will introduce additional labeling cost, but could make it easier to learn a robust imitator from a small training set. 

\item {\bf Model-based RL for safe learning:} Model-based agents could potentially learn which actions are catastrophic without ever trying them. They could achieve this by learning a good world model through exploration of safe regions of the state space. (Similarly, chemists know to avoid exposure to certain chemicals even if no human has ever been exposed to the chemical.)

\end{itemize}

\newpage
\section*{Acknowledgements}
This work was supported by Future of Life Institute grant 2015-144846 (all authors) and by the Future of Humanity Institute, Oxford. We thank Vlad Firoiu for early contributions and Jan Leike and David Abel for helpful comments. Special thanks to David Krueger for detailed comments on a draft. 
	
\bibliographystyle{plainnat}
\bibliography{arxiv}

\normalsize
\input{appendix}

\end{document}

%% file: appendix.tex
\newpage
\section{Appendix}
\label{section:appendix}

\subsection{Neural network architectures and hyperparameters}
\label{sec:neural}

\subsubsection{RL agent parameters
}

A3C agent network architecture (Pong, RoadRunner):

\begin{itemize}
\item Based on OpenAI's {\color{blue} \href{https://github.com/openai/universe-starter-agent}{Universe Starter Agent}}
\item Input format: 42x42x1, grayscale, (cropped, downsampled, rgb values averaged)
\item 4 convolutional layers with 32 3x3 filters, applied with 2x2 stride
\item Last convolutional layer fed into an LSTM with 256 hidden units
\item LSTM output fed into linear layers to produce value function estimate and policy logits
\item ELU activation
\item Learning rate: 0.0001
\item Adam Optimizer
\item Entropy bonus: 0.01
\item Discount factor: 0.99
\item Steps between policy gradient updates: 20
\end{itemize}

(Double) DQN agent network architecture (Space Invaders)
\begin{itemize}
\item Based on OpenAI's {\color{blue} \href{https://github.com/openai/baselines}{baseline DQN implementation}} using Double DQN
\item Input format: 84x84x1, grayscale, (cropped, downsampled)
\item Convolutional layer with 32 8x8 filters, 4x4 stride
\item Convolutional layer with 64 4x4 filters, 2x2 stride
\item Convolutional layer with 64 3x3 filters
\item Hidden layer with 512 units
\item Output layer
\item RELU activation
\item Adam Optimizer
\item Steps: 2500000
\item Exploration schedule: exploration rate is 1.0 until step 25000, then linearly decreased to 0.01 until step 1250000, then fixed at 0.01
\item Learning rate schedule: $10^{-4}$ until step 25000, linearly decreased to $5*10^{-5}$ until step 1250000, then fixed at $5*10^{-5}$
\item Gradient norm clipping: 10
\item Target network update frequency: 10000
\item Learning starts: 50000
\item Frame history length: 4
\item Replay buffer size: 1000000
\item Discount factor: 0.99
\item Batch size: 32
\item Frameskip: 4
\item Episode ended at end of life (but environment not reset until end of episode)
\end{itemize}

Game-dependent reward scaling
\begin{itemize}
\item Pong: reward = reward/1.0
\item Road Runner: reward = reward/100.0
\item Space Invaders: reward clipping to +/-1
\end{itemize}


\subsubsection{Blocker Parameters}

Parameters fixed across all experiments:

\begin{itemize}
\item Input format: [105, 80, 3], color (cropped then downsampled)
\item Convolutional layers, where final layer is concatenated with one-hot embedding of agent's action
\item FC layers and a linear layer outputting logits

\item Learning rate 0.002
\item Adam Optimizer
\item Batch size: 400
\end{itemize}

Pong:
\begin{itemize}
\item 2 convolutional layers, 4 filters size 3x3 with 2x2 stride
\item 2 10-unit hidden layers
\item No dropout
\end{itemize}

Space Invaders and Road Runner:
\begin{itemize}
\item 4 convolutional layers, 16 filters size 3x3 with 2x2 stride
\item 2 20-unit hidden layers
\item Dropout with probability of discarding 0.5
\item Examples were reweighted to give positive and negative examples equal weight
\item Labels were manually cleaned after collection (by manually reviewing episodes and by looking for individual frames where the blocker disagreed with the given label)
\end{itemize}

\subsection{How the Blocker Replaced Catastrophic Actions} \label{sec:replace}

The Blocker should be trained to not just imitate the human's classification of actions as catastrophic but also to decide which safe action to substitute for the catastrophic action (Fig 1). This would makes the supervised learning problem of training the Blocker more complex than just a binary classification task. In our experiments we avoid dealing with the more complex learning problem as it seems unlikely to change our conclusions. Instead, we use the following techniques:

\begin{itemize}
\item {\bf Fixed Action Replacement:}
The human specifies which action the Blocker should use to replace blocked actions. More generally, the human could specify a lookup table.

\item {\bf Action Pruning:}
If an action is blocked, it is not sent to the environment. The agent has to choose an action again (having received a penalty for the blocked action). To ensure the agent always has at least one action available, the action with the lowest logit score is never blocked. (Essentially, we wait until the agent chooses an action that the Blocker thinks is unlikely to be catastrophic. This is a technique for replacing actions that is learned rather than hard-coded by the human. But the more general strategy would be to learn to imitate how the human replaces actions.)
\end{itemize}

Here are the techniques used for each game:

\begin{itemize}
\item {\bf Pong:} Action Replacement with safe action ``Up''.

\item {\bf Space Invaders:} Action Replacement with the safe action being the agent's action but with ``Fire'' removed.

\item {\bf Road Runner:} Action Pruning. 

\end{itemize}




\subsection{Space Invaders Experiment: Human Oversight Procedure}
In Space Invaders, the agent starts on the left side of the screen. When a human blocks it from shooting the left barrier, it responds by staying to the left of the left barrier (where it knows it won't get a negative reward penalty). This means for that for many episodes it never goes under the middle or right barriers. To get a training set that includes shooting under those barriers, the human would have to label for a long time. (We estimate 70 hours.) 
We fixed this problem by including episodes where the agent is initially placed at the center or right of the screen. We alternated between episodes with these three different initializations (i.e.\ starting at left (as normal), starting at center, starting at right). Once the Human Oversight phase was complete, we reverted to the normal initialization for every episode (starting at left). 


